\documentclass[conference]{IEEEtran}
 
\IEEEoverridecommandlockouts
\usepackage{booktabs}
\usepackage{makecell}
\usepackage{tabularray}
\usepackage{cite}
\usepackage{amsmath,amssymb,amsfonts}
\usepackage{algorithmicx}
\usepackage{graphicx}
\usepackage{textcomp}
\usepackage{xcolor} 
\def\BibTeX{{\rm B\kern-.05em{\sc i\kern-.025em b}\kern-.08em
    T\kern-.1667em\lower.7ex\hbox{E}\kern-.125emX}}
\DeclareUnicodeCharacter{FF1A}{:}
\DeclareUnicodeCharacter{00B7}{\textemdash}
\DeclareUnicodeCharacter{FF08}{(}
\DeclareUnicodeCharacter{FF09}{)}
\usepackage{algorithm}
\usepackage{algpseudocode}

\begin{document}

\title{ Enhancing Multi-Agent Consensus through Third-Party LLM Integration: Analyzing Uncertainty and Mitigating Hallucinations in Large Language Models}

\author{\IEEEauthorblockN{Zhihua Duan}
\IEEEauthorblockA{\textit{Intelligent Cloud Network Monitoring Department  } \\
\textit{China Telecom Shanghai Company}\\
Shanghai, China \\
duanzh.sh@chinatelecom.cn}
\and
\IEEEauthorblockN{Jialin Wang$^*$}
\IEEEauthorblockA{\textit{Executive Vice President  } \\
\textit{Ferret Relationship Intelligence }\\
Burlingame, CA 94010, USA \\
jialinwangspace@gmail.com$^*$} 
}

\maketitle

\begin{abstract}
Large Language Models (LLMs) still face challenges when dealing with complex reasoning tasks, often resulting in hallucinations, which limit the practical application of LLMs. To alleviate this issue, this paper proposes a new method that integrates different LLMs to expand the knowledge boundary, reduce dependence on a single model, and promote in-depth debate among agents. The main contributions include: 1) Introducing third-party LLMs to adjust the attention weights of agents through uncertainty estimation and confidence analysis, optimizing consensus formation in multi-agent systems; 2) Experiments on arithmetic datasets have validated the effectiveness of the method, surpassing traditional multi-agent baselines. This research provides a new perspective for large models to alleviate hallucination phenomena when dealing with complex tasks.

\end{abstract}


\section{Introduction}
Large Language Models (LLMs) have exhibited robust reasoning and creative capabilities. In order to manage more intricate tasks, AutoGen amalgamates the functionalities of LLMs, human inputs, and various tools by constructing agents with distinct roles, thereby offering a tailored multi-agent dialogue setting\cite{AutoGen}.
This has catalyzed the evolution of multi-agent debate systems. In these systems, multiple agents articulate their arguments, while a neutral moderator oversees the debate process to facilitate the attainment of a final resolution\cite{Encouraging}.By employing multiple instances of language models and engaging in several rounds of proposal and debate regarding their respective answers and reasoning processes, a consensus final answer is ultimately achieved\cite{DebUnc}.

In the process of multi-agent debate, it is common to deploy the same Large Language Model (LLM) for each agent and assign them the same role. This approach has proven effective in multi-agent debates, but it also has limitations, especially when the model's knowledge scope is restricted and lacks external feedback from different models. The use of a single model can lead to a monolithic viewpoint in the debate process, limiting the depth and breadth of the debate.

\begin{figure}[htbp]
  \centering
  \includegraphics[width=\linewidth]{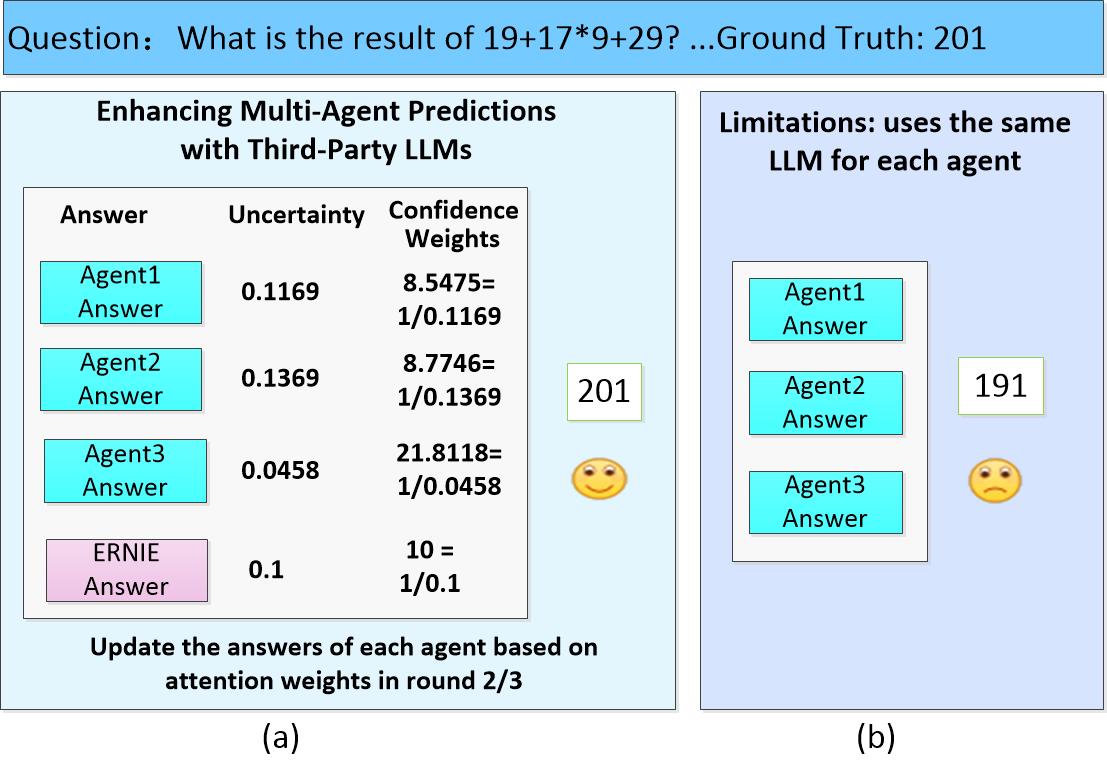}
  \caption{ As shown in Figure 1(a), we introduce third-party LLMs to enrich the knowledge of multi-agent system, which can provide agents with different perspectives. Agents can adjust their answers based on more information and deeper thinking. The uncertainty and confidence weighting mechanism allows each agent to dynamically adjust its attention weights in the second and third rounds of debate, so that it can self-adjust according to other agents’ feedback and approach the correct answer more accurately. As shown in Figure 1(b), all agents in traditional methods depend on a single LLM, which makes the answers have consistent bias and lack of diversity and depth of thinking.}
\end{figure}

Therefore, this paper raises a question: Can a third-party model be introduced to complete tasks through the collaboration of multiple Large Language Models? It explores combinations of different LLMs to estimate the confidence levels of each agent and seeks to achieve a broader consensus.

Our research motivation is to explore how the introduction of third-party Large Language Models (LLMs) and strategies for combining different LLMs can enhance the decision-making capabilities and problem-solving depth of multi-agent systems. Special emphasis is placed on pairing external LLMs with specific domain knowledge with existing LLMs; exploring different combinations of LLMs is deemed valuable to foster debate and collaboration among agents, thereby achieving a comprehensive and in-depth analysis of complex problems.
The main contributions of this paper are as follows:
1. This study introduces third-party Large Language Models (LLMs) to investigate the application of uncertainty estimation and confidence in multi-agent systems. By adjusting attention weights, the LLM model can learn from the responses of other agents, thereby facilitating the formation and optimization of consensus.
2. The research conducted in this paper involved experiments on an arithmetic dataset, demonstrating that the method proposed in this paper outperforms previous multi-agent baselines.
﻿

\section{Related Work}
\subsection{Uncertainty in LLMs}
While uncertainty in conventional natural language processing (NLP) tasks has been extensively researched, it provides a robust theoretical foundation and methodological guidance for harnessing uncertainty in NLP tasks\cite{Translation}\cite{DEUP}\cite{UncertaintyEstimation}.However, the field of uncertainty quantification in large models is still understudied. In generative AI large models, text is typically generated based on probabilistic models, considering different contexts and internal states during each generation. Any generated answer that is semantically consistent with the actual answer can be considered correct\cite{Multitask}\cite{Llama2}\cite{Mistral}.
Proposed to quantify uncertainty by directly prompting the language model to answer questions related to the uncertainty of its own generation\cite{Uncertainty }.

A new method for uncertainty quantification called Shifting Attention to more Relevant (SAR) has been proposed. This method focuses on more relevant components at both the token and sentence levels to achieve better uncertainty quantification\cite{ShiftingAttention}.
A multi-agent debate framework is proposed, which assesses the confidence level of agents through uncertainty measurement and adjusts the attention mechanism of the LLM to adjust token weights based on confidence levels. Text prompts are used to convey confidence, thereby enhancing reliability in multi-agent debates\cite{DebUnc}.

This study aims to introduce a third-party large model to specify uncertainty metrics, and to describe the uncertainty of generated tokens by transforming the logits values in the generation results.

\subsection{Applications of Agents in LLMs}
Recently, significant progress has been made in automated task resolution through the use of multi-agents driven by Large Language Models (LLMs).
ReConcile is a multi-model multi-agent framework that enhances collaborative reasoning between Large Language Models (LLMs) through multi-round discussions, employing a confidence-weighted voting mechanism to achieve better consensus\cite{ReConcile}.

By integrating large language models and tools, multiple agents can be introduced to facilitate mutual promotion of different roles, thereby enhancing reasoning capabilities through multi-agent debate\cite{AutoGen}\cite{MetaGPT}\cite{Autonomous}.
A smart personalized digital banking assistant based on LangGraph and thought chains has been proposed, utilizing Large Language Models (LLMs) and a multi-agent framework to improve task efficiency\cite{LangGraph}.

A multi-agent system approach has been adopted, implementing system 2 thinking through the CrewAI framework, which integrates in-depth analysis and reasoning decision-making to enhance AI's capabilities in handling complex problems\cite{CrewAI}.

This article constructs four agents and conducts three rounds of question-and-answer sessions, where agents share response information and uncertainty probabilities with each other to refine their answers, ultimately determining the final answer based on the principle of majority voting.

\section{Methods}
To accurately grasp multi-round dialogue information and enhance the reasoning capabilities of large models, as shown in Figure 2, this paper proposes a simple yet powerful fine-grained reasoning method. The entire dialogue session is divided into the context information of user questions and responses from four agent proxies, constructing a context prompt that includes the response information of four dialogue agents, and requiring the current agent to provide an updated answer based on the opinions of other agents, and to state the final answer, guiding the agent to consider all agents' dialogue and confidence levels comprehensively.

Attention weight update: In the open-source large model Llama, the Attention Scaling Range Weights mechanism is used, multiplying the range weights on the basis of attention, allocating more attention to the information of agents with larger range weights, thereby dynamically adjusting the large model's attention weights to different agents, and updating its output or decision.
In this way, the model can not only consider the contribution of each agent but also adjust the importance of each agent's dialogue according to confidence levels, thus generating a more comprehensive and balanced response. The system takes into account information from multiple agents when making decisions and can weight this information based on the credibility of each information source, thereby reaching a more accurate conclusion. After conversion through logits, it outputs text completion information and uncertainty measures.

\begin{figure*}[htbp]
  \centering
  \includegraphics[width=\linewidth]{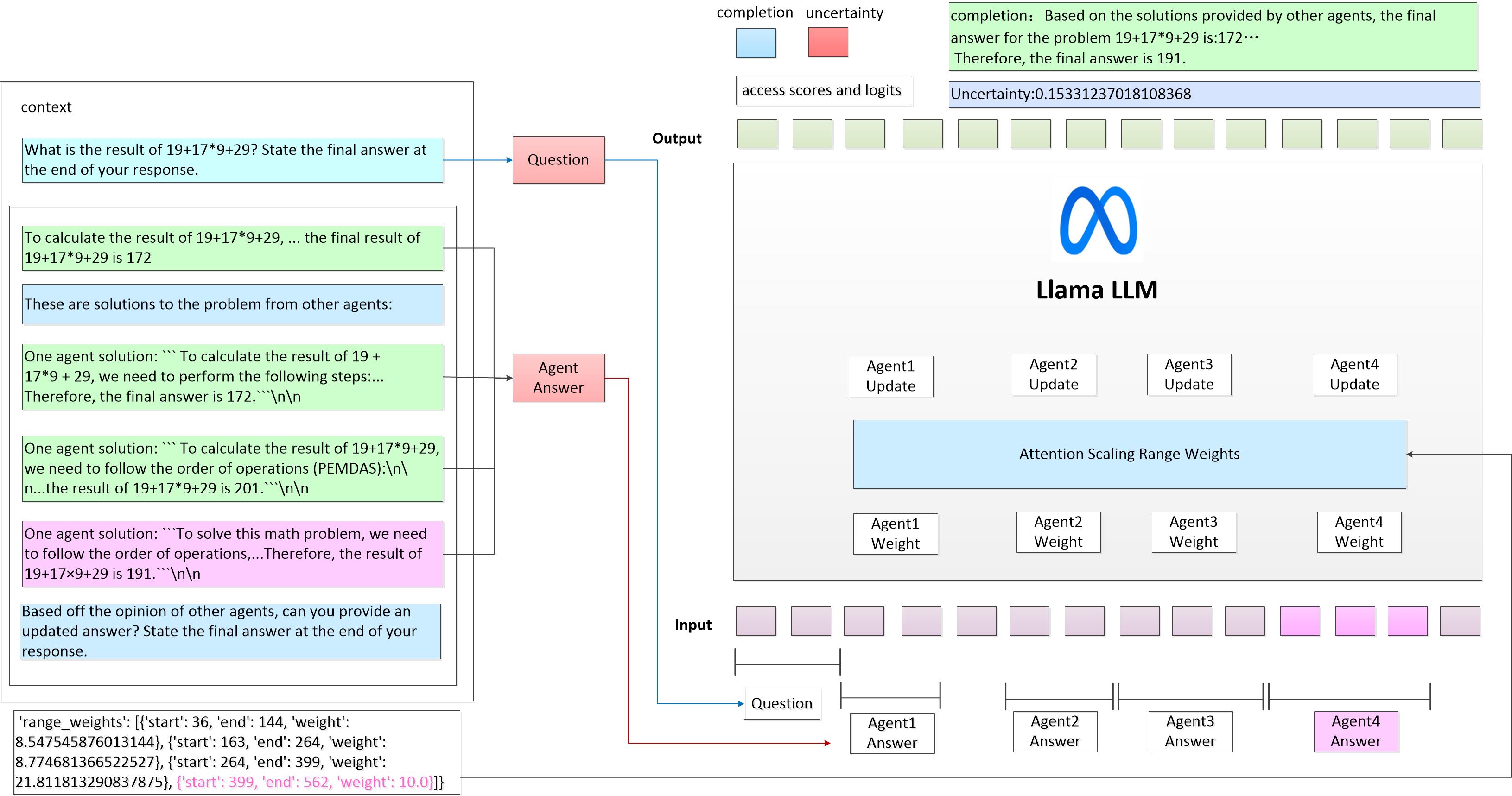}
  \caption{ The weight values of different agents represent their relative importance or trustworthiness when dealing with specific tasks. The range weights are a list that includes multiple weight items, each with a starting point and an ending point (start and end), as well as a corresponding weight value (weight). For example, when the fourth agent uses a third-party large language model to answer questions, its confidence parameter is set to 10.0, with the corresponding range being 'start': 399, 'end': 562, 'weight': 10.0. Within the range of processing context prompts, the large model will adjust the distribution of attention weights based on these parameters.Subsequently, the model outputs completion and its corresponding uncertainty value through the logits transformation mechanism. The confidence weight=1/uncertainty.For example, weight = 1/0.15331237018108368 = 6.522630879810011. In the second and third rounds of iteration, the agent's response and its confidence weight are used as reference information for other agents in subsequent rounds. Ultimately, a comprehensive response is derived after the second and third rounds of iteration.
}
   
\end{figure*}
In multi-agent dialogues, the model focuses on agents with higher confidence. After each round of dialogue, responses from each agent are received. Attention scaling is only applied to the answers from the previous round. For example, the second round scales attention weights based on the responses from the first round, and the third round scales attention weights based on the responses from the second round.
The large model uses Transformer decoder layers, where the attention mechanism creates "query," "key," and "value" vectors for each token. The similarity between the "query" vector of the current token and the "key" vectors of each token is used to calculate the weights for each token, which are normalized through the softmax function to ensure their sum is 1, and are used to create the output vector. The weight of each token determines its influence on the generation of the next token. By modifying these weights, the model can adjust its focus on each token in the input\cite{attention}.

The calculation formula for updating weights using range weights is as follows:

1. The original formula for calculating Transformer attention weights.\[ A = Q K^T \]
2. The improved formula applying weighted importance and range limitations.
\[ A[:, :, :, r_{\text{start}}: r_{\text{end}}] \leftarrow A[:, :, :, r_{\text{start}}: r_{\text{end}}] \cdot \lambda \cdot r_{\text{w}} \]
3. The calculation of attention weights after normalization and application of the softmax function.

\[ \text{Attention}(Q, K, V) = \text{softmax}\left(\frac{A}{\sqrt{d_k}}\right) V \]

A represents the attention weight matrix, \( Q \), \( K \) and \( V \) represent the Query, Key and Value matrices. \( d_{k} \) is the dimension of the key vectors. \( r_{\text{start}} \) and \( r_{\text{end}} \) represent the start and end positions of the range. \(\lambda\)  is a scalar used to adjust the importance weights, and \( r_{\text{w}} \) is a weight factor related to the range.

As depicted in Algorithm 1, the algorithmic framework employed in this study is based on the code implementation proposed in reference\cite{DebUnc}, with further optimizations applied. We provide a detailed description of the confidence-based attention adjustment mechanism within the LlamaAttention class. The process begins by evaluating the sum of attention weights across specified ranges, represented by range weights. For each range, the mean and standard deviation of the attention weights are calculated to derive a weighted importance measure. This measure, in conjunction with the range weights, is used to adjust the original attention weights, enhancing the impact of more confident predictions. Post-adjustment, a normalization step ensures that the adjusted weights maintain the same distribution as the original weights, thus preserving the overall attention distribution. This approach allows the model to focus more on areas with higher confidence, thereby improving the accuracy of the attention mechanism.

\begin{algorithm*}
\caption{Attention Weighting Mechanism Based on Confidence Adjustment}
\begin{algorithmic}[1]
\Procedure{LlamaAttention}{$Q, K, V, range\_weights$}
    \State $attn\_weights \gets \text{Compute attention weights}(Q, K)$
    \If{$range\_weights \neq \text{None} \land \text{len}(range\_weights) > 1 \land \text{attn\_weights}.shape[2] == 1$}
        \State $original\_sum \gets \sum_{rw \in range\_weights} \sum \text{attn\_weights}[:, :, :, rw.start : rw.end]$
        \For{$rw \in range\_weights$}
            \State $range\_data \gets \text{attn\_weights}[:, :, :, rw.start : rw.end]$
            \State $\mu \gets \text{mean}(range\_data, \text{dim}=-1, \text{keepdim}=True)$
            \State $\sigma \gets \text{std}(range\_data, \text{dim}=-1, \text{keepdim}=True)$
            \State $\text{weighted\_mean} \gets \mu \times rw.weight$
            \State $\text{weighted\_importance} \gets 1 + \frac{range\_data - \text{weighted\_mean}}{\sigma + 1e-5}$
            \State $\text{attn\_weights}[:, :, :, rw.start : rw.end] \gets \text{attn\_weights}[:, :, :, rw.start : rw.end] \times \text{weighted\_importance} \times rw.weight$
        \EndFor
        \State $new\_sum \gets \sum_{rw \in range\_weights} \sum \text{attn\_weights}[:, :, :, rw.start : rw.end]$
        \State $norm\_factor \gets \frac{original\_sum}{new\_sum}.unsqueeze(-1)$
        \For{$rw \in range\_weights$}
            \State $\text{attn\_weights}[:, :, :, rw.start : rw.end] \gets \text{attn\_weights}[:, :, :, rw.start : rw.end] \times norm\_factor$
        \EndFor
    \EndIf
     \State $\text{attn\_output} \gets \text{torch.matmul}(attn\_weights, \text{value\_states})$
    \State \Return $\text{attn\_output}, \text{attn\_weights}$   
\EndProcedure
\end{algorithmic}
\end{algorithm*}

\section{Experiment Design}

This experiment utilizes the Arithmetic dataset, which is a collection of randomly generated arithmetic problems in the mathematical form of \( a + b \times c + d \), where \( 0 \leq a, b, c, d < 30 \),As shown in Table 1. Considering the cost of running, only a dataset of 100 problems was constructed, and the uncertainty measures and methods for these samples were evaluated.  

\begin{table}
\centering
\caption{Arithmetic Problems and Their Solutions}
\begin{tblr}{
  width = \linewidth,
  colspec = {Q[40]Q[867]Q[27]},
  hline{1,3,5,7} = {1-2}{},
  hline{2,4,6} = {2}{},
}
Q: & What is the result of 3+27*3+7? State the final answer at the end of your response.   &  \\
A: & 91                                                                                    &  \\
Q: & What is the result of 9+19*21+18? State the final answer at the end of your response. &  \\
A: & 426                                                                                   &  \\
Q: & What is the result of 19+17*9+29? State the final answer at the end of your response. &  \\
A: & 201      &  \\
\end{tblr}
\end{table}

As shown in Figure 3, this experiment involves four agents and three rounds of question-and-answer, with the first three agents using the same LLM (Llama3\cite{Llama2}). The fourth agent introduces a third-party large model (ERNIE\cite{ERNIE}).

\begin{figure}[htbp]
  \centering
  \includegraphics[width=\linewidth]{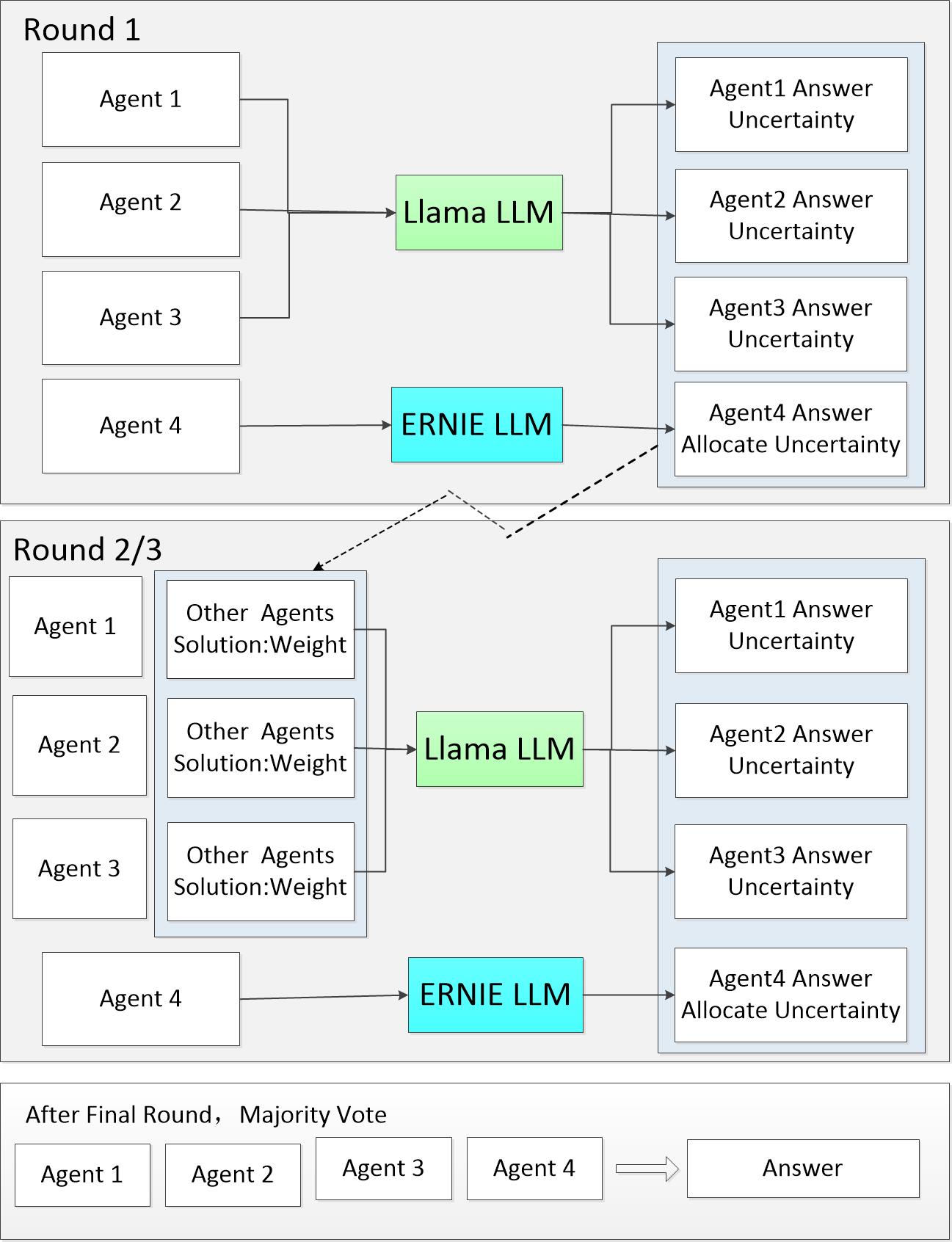}
  \caption{  In the first round, each agent answers the question individually. In the subsequent second and third rounds, each agent provides response information and uncertainty probabilities from other agents and uses this information to refine their answers. The first three agents use the Llama3 model, modify model weight information through attention scaling, output the model's predicted results, and calculate the uncertainty measure of this response through token probabilities. The fourth agent uses the ERNIE LLM model, specifying weight coefficients manually as uncertainty probabilities. The final answer is determined by a majority vote after the third round. }
   
\end{figure}

\section{Results} 
This study proposes a new method where the first three agents use the Llama large model, and the fourth agent introduces the third-party ERNIE large model. By employing uncertainty  and dynamically adjusting attention weights, the task performance is significantly improved after three rounds of dialogue. As shown in Table 2, the experimental results indicate that our method achieved significant effectiveness under the Attention-All setting, with an accuracy rate as high as 0.940, surpassing other baseline methods. Among them, the standard baseline method has an accuracy rate of 0.478, and the entropy-based baseline methods have accuracy rates of 0.482 and 0.518, respectively. The TokenSAR baseline method's accuracy rate ranges from 0.464 to 0.500. The Oracle method's accuracy rates are 0.542, 0.654, and 0.732.

\begin{table}
\centering
\caption{ Performance comparison of different methods}
\begin{tblr}{
  width = \linewidth,
  colspec = {Q[273]Q[323]Q[287]},
  hline{1-3,6,9,12-13} = {-}{},
  hline{4-5,7-8,10-11} = {2-3}{},
}
Estimator & Method      & Arithmetic \\
N/A$^{\dag}$       & Standard    & 0.478      \\
          & Prompt      & 0.482      \\
Entropy$^{\dag}$   & Attn-Others & 0.518      \\
          & Attn-All    & 0.518      \\
          & Prompt      & 0.464      \\
TokenSAR$^{\dag}$  & Attn-Others & 0.500      \\
          & Attn-All    & 0.500      \\
          & Prompt      & 0.542      \\
Oracle$^{\dag}$    & Attn-Others & 0.654      \\
          & Attn-All    & 0.732      \\
ours      & Attn-All    & 0.940      
\end{tblr}
\footnotesize{ 

Scores with $^{\dag}$ were obtained from the paper\cite{DebUnc}}\\
\end{table}

\section{Limitations}
This study demonstrates potential in specific tasks, but there are limitations in computational efficiency, application of attention mechanisms, and cross-domain experiments. Future research needs to address these issues to expand the application scope of LLMs.

1. Computational Overhead and Real-time Application Challenges: This method enhances the ability to handle complex tasks by integrating third-party large models, but this correspondingly increases the computational cost. Therefore, when implementing a multi-agent architecture, it is necessary to consider the potential impact on computational efficiency and make corresponding trade-offs.

2. Application of Attention Weighting Mechanism: Future research can further explore the applicability of this mechanism in diverse tasks and how to adjust the attention weights according to the characteristics of different tasks to enhance the model's overall performance.

3. Extensive Experiments in Specific Domains: This study has achieved certain results in model experiments in specific domains, but the scope of the experiments is limited. Although the performance of this method on datasets such as MMLU and TruthfulQA has not yet reached the performance level of current state-of-the-art (SOTA) large models, there is potential for future improvement by employing more powerful large models.

\section {Conclusion}
This paper effectively addresses the trust issue when multiple agents provide different answers by introducing a third-party Large Language Model (LLM) and integrating its responses and confidence levels into multi-agent dialogues. To calculate confidence, this paper optimizes the attention mechanism of the LLM by setting confidence parameters for the third-party LLM and combining them with the confidence of the Llama model, achieving attention weight adjustment based on confidence levels. Experiments show that this method can more effectively convey information to the LLM compared to traditional text prompts, allowing the model to consider the contributions and confidence levels of each agent when generating responses, thus producing more comprehensive and balanced outputs. This study not only enhances the dialogue system's ability to handle uncertainty but also provides a new perspective for the future application of LLMs in multi-agent interactions. Future work will explore how to further apply the attention scaling mechanism to a broader range of scenarios.

 

\begin{thebibliography}{10}

\bibitem{AutoGen}
Qingyun Wu, Gagan Bansal, et~al.
\newblock Autogen: Enabling next-gen llm applications via multi-agent conversation framework.
\newblock {\em CoRR}, abs/2308.08155, 2023.

\bibitem{Encouraging}
Tian Liang, Zhiwei He, et~al.
\newblock Encouraging divergent thinking in large language models through multi-agent debate.
\newblock {\em CoRR}, abs/2305.19118, 2023.

\bibitem{DebUnc}
Luke Yoffe, Alfonso Amayuelas, et~al.
\newblock Debunc: Mitigating hallucinations in large language model agent communication with uncertainty estimations.
\newblock {\em CoRR}, abs/2407.06426, 2024.

\bibitem{Translation}
Marina Fomicheva, Shuo Sun, et~al.
\newblock Unsupervised quality estimation for neural machine translation.
\newblock {\em Intelligent Systems in Accounting, Finance Management}, 8:539--555, 2020.

\bibitem{DEUP}
Salem Lahlou, Moksh Jain, et~al.
\newblock Deup: Direct epistemic uncertainty prediction.
\newblock {\em Computing Research Repository}, abs/2102.08501, 2021.

\bibitem{UncertaintyEstimation}
Yuxia Wang, Daniel Beck, et~al.
\newblock Uncertainty estimation and reduction of pre-trained models for text regression.
\newblock {\em Transactions of the Association for Computational Linguistics}, 10:680--696, 2022.

\bibitem{Multitask}
Alec Radford, Jeffrey Wu, et~al.
\newblock Language models are unsupervised multitask learners.
\newblock {\em OpenAI blog}, 1(8):9, 2019.

\bibitem{Llama2}
Hugo Touvron, Louis Martin, et~al.
\newblock Llama 2: Open foundation and fine-tuned chat models.
\newblock {\em CoRR}, 2023.

\bibitem{Mistral}
Albert~Q. Jiang, Alexandre Sablayrolles, et~al.
\newblock Mistral 7b.
\newblock {\em CoRR}, abs/2310.06825, 2023.

\bibitem{Uncertainty}
Stephanie Lin, Jacob Hilton, et~al.
\newblock Teaching models to express their uncertainty in words.
\newblock {\em Trans Mach Learn Res}, 2022.

\bibitem{ShiftingAttention}
Jinhao Duan, Hao Cheng, et~al.
\newblock Shifting attention to relevance: Towards the predictive uncertainty quantification of free-form large language models.
\newblock {\em Proceedings of the 62nd Annual Meeting of the Association for Computational Linguistics}, pages 5050--5063, 2024.

\bibitem{ReConcile}
Justin Chih-Yao Chen, Swarnadeep Saha, et~al.
\newblock Reconcile: Round-table conference improves reasoning via consensus among diverse llms.
\newblock {\em Proceedings of the 62nd Annual Meeting of the Association for Computational Linguistics}, abs/2309.13007:7066--7085, 2024.

\bibitem{MetaGPT}
Sirui Hong, Mingchen Zhuge, et~al.
\newblock Metagpt: Meta programming for multi-agent collaborative framework.
\newblock {\em International Conference on Learning Representations}, 2024.

\bibitem{Autonomous}
Lei Wang, Chen Ma, et~al.
\newblock A survey on large language model based autonomous agents.
\newblock {\em Frontiers of Computer Science}, 18(6):186345, 2024.

\bibitem{LangGraph}
Arafat~Md Easin, Saha Sourav, et~al.
\newblock An intelligent llm-powered personalized assistant for digital banking using langgraph and chain of thoughts.
\newblock {\em IEEE 22nd Jubilee International Symposium on Intelligent Systems Informatics}, pages 625--630, 2024.

\bibitem{CrewAI}
P.~Venkadesh, S.~V. Divya, et~al.
\newblock Unlocking ai creativity: A multi-agent approach with crewai.
\newblock {\em Journal of Trends in Computer Science Smart Technology}, 6(4):338--356, 2024.

\bibitem{attention}
Ashish Vaswani, Noam Shazeer, et~al.
\newblock Attention is all you need.
\newblock {\em Advances in neural information processing systems}, 30:5998--6008, 2017.

\bibitem{ERNIE}
Yu~Sun, Shuohuan Wang, et~al.
\newblock Ernie 2.0: A continual pre-training framework for language understanding.
\newblock {\em Proceedings of the AAAI Conference on Artificial Intelligence}, 34(05):8968--8975, 2020.

\end{thebibliography}

\end{document}